\documentclass[runningheads]{llncs}
\usepackage[T1]{fontenc}
\usepackage{times}
\usepackage[hyphens]{url}
\usepackage{algorithm}
\usepackage{algorithmic}
\usepackage{newfloat}
\usepackage{listings}
\usepackage{amsmath}
\usepackage{graphicx}
\usepackage{xspace}
\usepackage{xcolor}
\usepackage[misc]{ifsym}
\newcommand{\our}{\text{RANA}\xspace}
\usepackage{subfig}
\begin{document}
\title{Relation-Aware Network with Attention-Based Loss for Few-Shot Knowledge Graph Completion}
\titlerunning{\our for Few-Shot Knowledge Graph Completion}
%
\author{Qiao Qiao \and
Yuepei Li \and
Kang Zhou 
\and
Qi Li}
%
%
\institute{Iowa State University, Ames, Iowa, USA
\email{\{qqiao1,liyp0095,kangzhou,qli\}@iastate.edu}}
%
\maketitle              
\begin{abstract}
Few-shot knowledge graph completion (FKGC) task aims to predict unseen facts of a relation with few-shot reference entity pairs.  Current approaches randomly select one negative sample for each reference entity pair to minimize a margin-based ranking loss, which easily leads to a zero-loss problem if the negative sample is far away from the positive sample and then out of the margin. Moreover, the entity should have a different representation under a different context. To tackle these issues, we propose a novel Relation-Aware Network with Attention-Based Loss (\our) framework. Specifically, to better utilize the plentiful negative samples and alleviate the zero-loss issue, we strategically select relevant negative samples and design an attention-based loss function to further differentiate the importance of each negative sample. The intuition is that negative samples more similar to positive samples will contribute more to the model. Further, we design a dynamic relation-aware entity encoder for learning a context-dependent entity representation. Experiments demonstrate that \our outperforms the state-of-the-art models on two benchmark datasets. 

\end{abstract}

\section{Introduction}\label{sec:intro}

Knowledge graphs (KGs) contain rich triples (facts), where each triple $(h, r, t)$ illustrates a relation $r$ between a head entity $h$ and a tail entity $t$. KGs such as Wikidata \cite{vrandevcic2014wikidata} and NELL \cite{carlson2010toward} have been applied to various downstream applications such as relation extraction \cite{zhou2023improving}, named entity recognition \cite{zhou2022distantly}, and node classification \cite{rong2019dropedge}. 

Knowledge Graph Completion (KGC) is proposed to solve the issue of incompleteness caused by missing entities or relations in the KGs. KG embedding \cite{bordes2013translating,trouillon2016complex} has achieved considerable performance on KGC. These models perform well with enough training triples, but a large portion of relations in KGs follow a long-tail distribution. For example, around 10\% of relations in Wikidata \cite{chen2019meta} have no more than 10 triples. Relations that do not have enough training triples are known as few-shot relations. It is crucial and challenging for the model to predict relations with limited training triples. 

Few-shot knowledge graph completion (FKGC) methods have been proposed to address the few-shot relation issue. Given the relation $r$ and few-shot reference entity pairs $(h,t)$, the FKGC aims to rank candidate tail entities $t$ for each query $(h,?)$. These few-shot reference entity pairs form a support set, and queries form a query set. One line of the existing methods focuses on designing metric learning algorithms to compute the similarity between entity pairs \cite{xiong2018one,zhang2020few}. Another line leverages model agnostic meta-learning algorithm (MAML) \cite{finn2017model} to learn the optimal parameters of the model \cite{chen2019meta,niu2021relational,wu2022hierarchical}. 

To train the model, current FKGC methods apply a margin-based ranking loss function that aims to separate the score of the positive triple from the score of the negative triple by a margin. One negative triple is formed for each positive triple by replacing the true tail entity with a randomly selected candidate tail entity. This loss function does not effectively utilize the negative samples. Furthermore, an irrelevant negative sample is likely to be selected due to a large number of candidates. These irrelevant negative samples lead to zero loss because the negative triple is far away from the positive triple. Therefore these irrelevant negative samples would not contribute to the training and slow down the convergence rate \cite{schroff2015facenet}. For example, given a true triple \textit{(Kobe Bryant, WorkIn, California)}, the model can select negative tail entities, such as \textit{New York, Thailand, London}, etc. Because \textit{Thailand} is irrelevant to the true tail entity \textit{California}, the distance between \textit{California} and \textit{Thailand} is greater than a predefined margin, and the corresponding loss is zero. Thus \textit{Thailand} may not contribute to the training.

To address the above limitations, we propose a framework called \our (\textbf{R}elation-\textbf{A}ware \textbf{N}etwork with \textbf{A}ttention-Based Loss). To improve the quality of negative samples, we propose to filter irrelevant candidate tail entities first and then randomly sample multiple negative samples instead of one. Since the importance of negative samples is different and depends on their similarities to the positive sample, we apply an attention mechanism to assign a weight to each negative sample, where the weights of the most relevant negative samples are higher than the weights of the less relevant negative samples. The attention-based weighted loss function can enable the model to effectively avoid zero-loss issues and thus learn a better decision boundary.

Further, we propose a context-dependent dynamic relation-aware entity encoder to learn different representations of an entity in different relations. Specifically, given a target relation and its support set, the entity encoder uses the similarities between the target relation and neighboring relations to differentiate the impact of neighboring entities and dynamically encode the local connections of the entity.
Finally, \our employs meta-learning to enable the model to perform well on a new relation with a few training triples in a small number of gradient steps. 

In summary, our main contributions are:
\begin{enumerate}
    \item We propose a new negative sampling strategy and a novel attention-based loss function to solve the zero-loss and slower convergence issues.
    \item We propose a dynamic entity encoder to learn a context-dependent entity representation and reduce the influence of unrelated neighboring entities.
    \item Experiment results on benchmark datasets show that \our consistently and significantly outperforms other baseline methods.
\end{enumerate}

\section{Related Work}\label{sec:related}
\subsection{Embedding based Knowledge Graph Completion}
Knowledge graph embedding aims to embed entities and relations into a low-dimensional continuous vector space while preserve their semantic meaning. 
Existing methods can be divided into the following categories: (1) Translation-based models calculate the Euclidean distance between entities and relations as the plausibility of a fact, such as TransE \cite{bordes2013translating}, RotatE \cite{sun2018rotate}, and TransMS \cite{yang2019transms}; 
(2) Semantic matching-based models calculate the semantic similarity between entities and relations as the plausibility of a fact, such as RESCAL \cite{nickel2011three}, DistMult \cite{yang2015embedding}, and PUDA \cite{tang2022positive}; and (3) Neural network-based models take entities and relations into a deep neural network to fuse the graph network structure and content information of entities and relations, such as SME \cite{bordes2014semantic}, CompLEx \cite{trouillon2016complex}, and BertRL \cite{zha2022inductive}. 
All above models require sufficient training triples and thus impair their performance on few-shot relations.

\subsection{Few-Shot Knowledge Graph Completion}
FKGC requires the model to predict new facts with a few training facts. Existing methods fall into two categories: (1) Metric-based models aim to learn the matching metrics by calculating the similarity between the query set and the support set. GMathching \cite{xiong2018one} focuses on one-shot KGC by considering both the learned embeddings and local graph structures. FSRL \cite{zhang2020few} and FAAN \cite{sheng2020adaptive} extend GMatching to few-shot scenarios. (2) Optimization-based models aim to learn a set of good initial model parameters so that the learned model can be generalized to the new relation quickly. MetaR \cite{chen2019meta}, GANA \cite{niu2021relational}, and HiRe \cite{wu2022hierarchical} focus on extracting relation-specific meta information from the embeddings of entities in the support set and transferring it to the query set. 

However, all these methods use a margin-based ranking loss, which can not effectively avoid the low-quality negative sample, leading to a zero-loss issue and influencing the convergence rate. Negative sampling has been proven as important as positive sampling in determining the optimization objective \cite{yang2020understanding}. Especially under the few-shot setting, given limited positive samples, how to select high-quality negative samples based on the corresponding positive sample is crucial.

\section{Preliminary}\label{sec:preliminary}

\subsection{Problem Definition}
\noindent \textbf{Knowledge Graph $\mathcal{G}$.}
A knowledge graph $\mathcal{G}$ is a set of triples $\mathcal{T} = \{(h,r,t) \subseteq \mathcal{E} \times \mathcal{R} \times \mathcal{E}\}$, where $\mathcal{E}$ and $\mathcal{R}$ represent the entity set and relation set, respectively. The relation set $\mathcal{R}$ contains few-shot relations and high-frequency relations. The background knowledge graph $\mathcal{G}_{background}$ is a set of triples associated with all high-frequency relations.

\noindent \textbf{Knowledge Graph Completion.}
The KGC task is to either predict the tail entity $t$ given the head entity $h$ and the query relation $r$: $(h,r,?)$ or predict unseen relation $r$ between two existing entities: $(h,?,t)$. In this work, we focus on tail entity prediction.

\noindent \textbf{Few-shot Knowledge Graph Completion.}
Given a relation $r \in \mathcal{R}$ and its few-shot support set $\mathcal{S} = \{(h_i,t_i) \in \mathcal{T}\}$, the FKGC task aims to predict tail entity $t$ for each query $\mathcal{Q} = \{(h_i,?) \in \mathcal{T}\}$. 

\noindent \textbf{A Few-shot Relation's Neighborhood.}
Given a triple $(h,r,t)$ of a few-shot relation $r$, the neighborhood of $r$ is defined as $\{h,t,\mathcal{N}_h,\mathcal{N}_t\}$, where $\mathcal{N}_h$ and $\mathcal{N}_t$ are the sets of one-hop 
neighbors of $h,t$, respectively. All $\mathcal{N}_h$ and $\mathcal{N}_t$ are from the background knowledge graph $\mathcal{G}_{background}$. A neighbor in $\mathcal{N}_h$ or $\mathcal{N}_t$ is composed of a neighboring relation $r_i$ and a neighboring entity $c_i$. We denote the neighbor of each entity ($h$ or $t$) 
as $\mathcal{N}_e = \{(r_i,c_i)|(e,r_i,c_i) \in \mathcal{G}_{background} \}$.

\subsection{Meta-learning Settings}

Meta-learning aims to train a model on several related tasks so that the model can quickly learn a new task using a few training data. We leverage an optimization-based meta-learning algorithm called MAML \cite{finn2017model}, which aims to learn a task-specific parameter set $\mathbf{\Theta}_i$ by using well-initialized meta-model parameter set $\mathbf{\Theta}$. It can be divided into two stages, meta-training and meta-testing. During meta-training, given a task $\mathcal{T}_i$, a support set $\mathcal{S}_i$ and a query set $\mathcal{Q}_i$ are first sampled from $\mathcal{T}_i$. Then, the model learns a task-specific parameter set $\mathbf{\Theta}_i$ by one gradient descent update on the support set $\mathcal{S}_i$:
\begin{equation}
    \mathbf{\Theta}_i = \mathbf{\Theta} - \eta * \mathbf{\nabla\mathcal{L}_{\mathcal{S}_i}}(\mathbf{\Theta}).
\end{equation}

Finally, meta-optimization across all query sets of tasks is performed to learn the meta-model parameter set $\mathbf{\Theta}$ by using task-specific parameter set $\mathbf{\Theta}_i$. During meta-testing, the model can quickly adapt to a new task using only a support set $\mathcal{S}$.

In FKGC, each task is defined as predicting new triples for a specific few-shot relation. All the relations in the meta-training form a meta-training set $\mathcal{R}_{meta-training}$. Since the goal is to predict facts of unseen relations, the relations in meta-validation $\mathcal{R}_{meta-validation}$, meta-testing $\mathcal{R}_{meta-testing}$, and $\mathcal{R}_{meta-training}$ are distinct.

\section{Methodology}\label{sec:methodology}
\begin{figure*}[!t]
\centering
\includegraphics[scale=0.085]{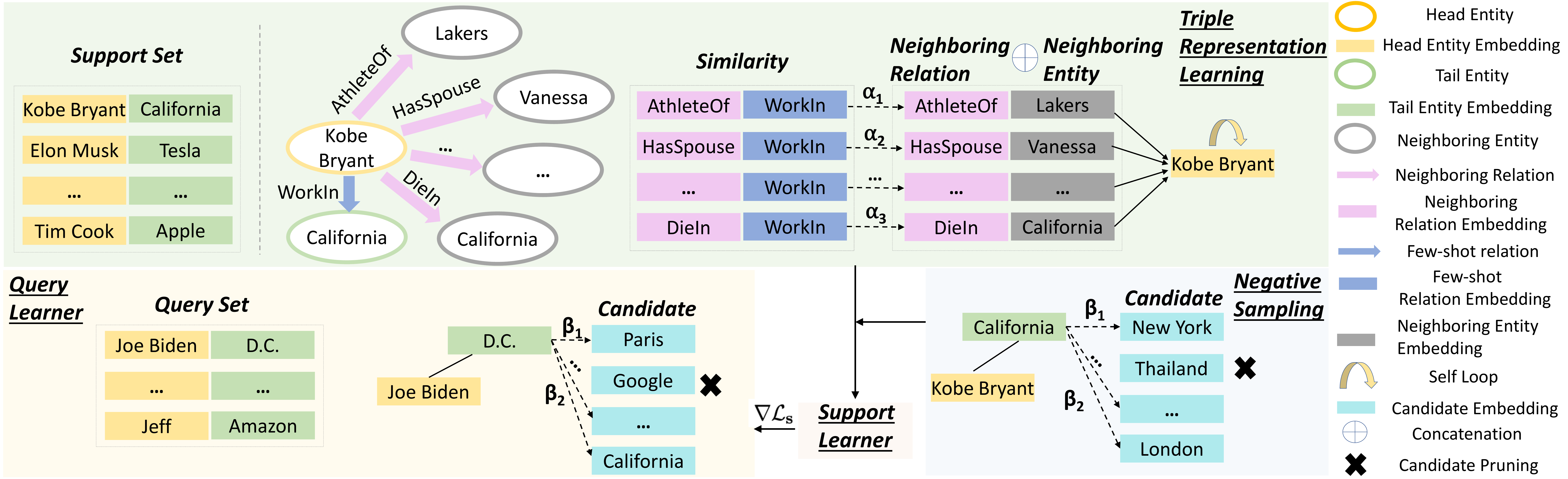}
\caption{The framework of \our for a few-shot relation \textit{WorkIn}}
\label{fig:1}
\end{figure*}

In this section, we first introduce triple representation, which aims to learn a context-dependent entity representation and a good initialization few-shot relation representation. Then we introduce a novel negative sampling strategy, which aims to filter irrelevant candidate tail entities and use an attention mechanism to differentiate the importance of each negative sample. Finally, we introduce meta-learning, which aims to learn well-generalized parameters so that the model can quickly adapt to a new task using few reference triples. Fig.\ref{fig:1} shows the framework of \our for a few-shot relation \textit{WorkIn}.

\subsection{Triple Representation}

\noindent \textbf{Dynamic Relation-Aware Entity Encoder.} 
The entity representation should be context-dependent. For example, \textit{(Kobe Bryant, California)} can involve in two different relations, such as \textit{WorkIn} and \textit{DieIn}, so \textit{Kobe Bryant} should have different embeddings in these two different relation contexts.

Besides, given a few-shot relation, different neighbors should have a different impact on the entity itself. For example, in Fig.\ref{fig:1}, given the few-shot relation \textit{WorkIn} and the head entity \textit{Kobe Bryant}, its neighbor \textit{(AthleteOf, Lakers)} should get more attention since it reveals work information about \textit{Kobe Bryant}, but the neighbor \textit{(HasSpouse, Vanessa)} should get less attention since it reveals family information of \textit{Kobe Bryant} which is irrelevant to the few-shot relation \textit{WorkIn}.

To address these issues, we design a dynamic relation-aware entity encoder, which incorporates neighboring relations to learn different embeddings of an entity in different relations and differentiates the importance of each neighbor by an attention mechanism. 

Given an entity pair $(h,t)$ from a support set $\mathcal{S}$, 
the embedding of few-shot relation $r$ is defined as:
\begin{equation}\label{eq:1}
    \mathbf{r = t - h},
\end{equation}
where $\mathbf{h}$ and $\mathbf{t}$ are the pretrained embeddings by TransE \cite{bordes2013translating}.

Here, we use the head entity $h$ as an example to illustrate the entity encoding procedure, and this procedure also holds for the tail entity $t$.

To differentiate the impact of each neighbor, we use a Multilayer Perceptron (MLP) network to calculate the relevance score between the few-shot relation $r$ and each neighboring relation $r_i$. 

The relevance score is defined as follows:
\begin{equation}\label{eq:2}
    m(r,r_i)=\mathbf{W_2[tanh(W_1[r \oplus r_i])]},
  \end{equation}

where $\oplus$ denotes the concatenation operation, $\mathbf{r_i}$ denotes the embedding of neighboring relations, and $\mathbf{W_1}$ and $\mathbf{W_2}$ are trainable parameters. A higher relevance score between the neighboring relation and the few-shot relation means that this neighboring relation is more important to the few-shot relation.

To learn the different representations of an entity in different relations, we design a dynamic neighbor embedding $\mathbf{A_{r_i,c_i}}$ of the head entity $h$ as follows:
\begin{equation}\label{eq:3}
  \mathbf{A_{r_i,c_i}} = \sum_{(r_i, c_i) \in \mathcal{N}_h} \alpha_i \mathbf{W_3[r_i \oplus c_i]},
  \end{equation} 
where $\mathbf{W_3}$ are trainable parameters, and $\alpha_i$ is the attention score of each neighbor: 
\begin{equation}\label{eq:4}
      \alpha_i = \frac{exp(m(r,r_i))}{\sum_{r_i \in \mathcal{N}_h} exp(m(r,r_i))}.
  \end{equation}
When the neighboring relation is more relevant to the few-shot relation, the higher attention $\alpha_i$ is given to the corresponding neighbor. Then this neighbor will play a more important role in neighbor embedding.

Since the information of entity $h$ itself is still valuable, we combine the embedding of entity $h$ with $\mathbf{A_{r_i,c_i}}$ to get the final representation $\mathbf{h'}$ as follows:
\begin{equation}\label{eq:5}
    \mathbf{h'} = \sigma \mathbf{(W_4(h + A_{r_i,c_i}))},
\end{equation}
where $\mathbf{W_4}$ are trainable parameters and $\sigma(\cdot)$ is a sigmoid function.
  
\noindent \textbf{Few-Shot Relation Representation.}
The same entity pair may involve in different relations, so the learning of relation representation is necessary, and it can further help triple representation learning. 

The relation representation from a specific entity pair in the support set $\mathcal{S}$ is:
\begin{equation}\label{eq:6}
\mathbf{R_{(h_i,t_i)}} = FC^{\sigma}_{\mathbf{W_5}}\mathbf{[h'_i \oplus t'_i]},
\end{equation}
where the fully connected layer $FC^{\sigma}_{\mathbf{W_5}}$ is parameterized by $\mathbf{W_5}$ and activated by a LeakyReLU function $\sigma(\cdot)$.

The relation representation from the support set $\mathbf{R^s}$ is then the average of all representations from entity pairs in $\mathcal{S}$,
\begin{equation}\label{eq:7}
    \mathbf{R^s} = \frac{\sum_{i=1}^I\mathbf{R_{(h_i,t_i)}}}{I}, 
\end{equation}
where $I$ is the number of entity pairs in the support set $\mathcal{S}$.

\subsection{Negative Sampling}

Since the positive sample is limited under the few-shot setting, how to take advantage of  negative samples is more critical. Previous FKGC methods use a margin-based ranking loss and randomly select one negative sample for each positive sample \cite{xiong2018one,chen2019meta,zhang2020few,sheng2020adaptive,niu2021relational}. But the random selection is likely to select an irrelevant negative sample and lead to a zero-loss issue. Further, regardless of their relevance to the positive samples, all negative samples will have the same impact on the model training. To address these issues, \our filters irrelevant negative samples and uses an attention mechanism to distinguish the importance of each negative sample.

\noindent \textbf{Candidate Pruning.}
The candidate set of negative samples constructed by \cite{xiong2018one} limits the candidate entities to those have the same types as the true tail entities in the support set, but this broad candidate set includes many irrelevant candidates as negative samples. For example, given a fact \textit{(Kobe Bryant, WorkIn, California)}, the previous candidate set is limited to location and company types of entities because the types of tail entities in the support set are company or location. However, a candidate such as \textit{Thailand} is irrelevant to \textit{California} and thus is not helpful in the model training. 

To reduce the number of irrelevant candidates and enable the model to select high-quality negative samples during the training stage, \our filters irrelevant candidates by the similarity of the true tail entity $t$ and a candidate tail entity $t^-$. The similarity is calculated by:
\begin{equation}\label{eq:8}
f(\mathbf{t,t^-}) = \mathbf{t^{-T}t},
\end{equation}
where $\mathbf{t}$ is the embedding of a true tail entity and $\mathbf{t^-}$ is the embedding of a candidate tail entity. If $f(\mathbf{t,t^-}) < \tau$, where $\tau$ is a threshold, then $t^-$ should be filtered. 

\noindent \textbf{Attention of Negative Samples.}
To fully utilize the negative samples, \our selects multiple negative samples instead of one and differentiates each negative triple's contribution by an attention mechanism.

Intuitively, if a negative sample is more relevant to the positive sample, this negative sample should play a more important role in model training. Therefore, higher attention should be given to this negative sample. As shown in Fig.\ref{fig:1}, given a positive sample \textit{(Kobe Bryant, California)}, the negative sample \textit{(Kobe Bryant, New York)} is more relevant to the positive sample than the negative sample \textit{(Kobe Bryant, London)}, and thus the model should pay more attention to the former. 

We define a scaled-dot product function $f(\mathbf{p_i,n_{ij}})$ to calculate the similarity between the positive sample $(h_i,t_i)$ and each of its negative sample $(h_i,t_{ij}^-)$:
\begin{equation}\label{eq:9}
\begin{gathered}
     \mathbf{p_i = h_i\oplus t_i}, \quad\quad \mathbf{n_{ij} = h_i\oplus t_{ij}^-},\quad\quad
    f(\mathbf{p_i,n_{ij}}) = \frac{\mathbf{n_{ij}^T p_i}}{\sqrt{|p|}},
\end{gathered}
\end{equation}
where $|p|$ is the dimension of $\mathbf{p_i}$.
The attention of each negative triple is defined by:
\begin{equation}\label{eq:10}
    \beta_{ij}=\frac{exp\, f(\mathbf{p_i,n_{ij}})}{\sum_{j=1}^J exp\, f(\mathbf{p_i,n_{ij})}},
\end{equation}
where $J$ is the number of negative samples.

\noindent \textbf{The Loss of \our.}
Negative sampling is as valuable as positive sampling in determining the optimization object, but it has been overlooked in the margin-based ranking loss \cite{yang2020understanding}. To alleviate zero-loss and slower convergence issue, we sample multiple negative triples instead of one to increase the probability of generating a relevant negative triple. 

Motivated by TransE \cite{bordes2013translating}, we first calculate the distance of each entity pair $(h_i,t_i)$ as follows:
\begin{equation}\label{eq:11}
    d_{(h_i,t_i)} = || \mathbf{h_i+R-t_i} ||_{L2},
\end{equation}

Because the smaller distance indicates the triple is more likely to be true, the triple should lead to a higher score. 
The score function of each triple is designed as:
\begin{equation}\label{eq:12}
    s_{(h_i,t_i)} = \gamma - d_{(h_i,t_i)},
\end{equation}
where $\gamma$ is a hyperparameter.

Our log-based loss function is:
\begin{equation}\label{eq:13}
 \mathcal{L} = -\sum_{i=1}^I log\, \sigma(s_{(h_i,t_i)})
-\sum_{i=1}^I\sum_{j=1}^J \beta_{ij} log\, \sigma(-s_{(h_i,t^-_{ij})}),   
\end{equation}
where $\sigma(\cdot)$ is a sigmoid function, and $\beta_{ij}$ is the attention of each negative triple calculated by Eq.\eqref{eq:10}. Since a more relevant negative triple has higher attention ($\beta_{ij}$), this loss function will make those relevant negative triples impact more in model training.

\subsection{Meta Learning}

To learn a new relation quickly with a support set, \our employs MAML \cite{finn2017model} to optimize the model parameters that can be adapted for few-shot relations. 

\noindent \textbf{Support Learner.} Support learner aims to learn a representation $\mathbf{R^s}$ of the few-shot relation and $\mathbf{R^s}$ can be calculated by Eq.\eqref{eq:1}-Eq.\eqref{eq:7}.

\noindent \textbf{Query Learner.} Following the MetaR \cite{chen2019meta} assumption, the relation is the key common information between support and query set. So we aim to transfer the support relation $R^s$ to the query relation $R^q$ by minimizing a loss function via gradient descending.

In \our, the relation embedding $\mathbf{R^q}$ can be updated by the gradient descent,
\begin{equation}\label{eq:14}
    \mathbf{R^q} = \mathbf{R^s} - \eta * \mathbf{\nabla\mathcal{L}_s},
\end{equation}
where the hyperparameter $\eta$ refers to the step size and $\mathcal{L}_s$ refers to the loss of the corresponding support set, which is calculated by Eq.\eqref{eq:11}-Eq.\eqref{eq:13}.

To update all parameters of \our, we use the updated relation embedding $\mathbf{R^q}$ to calculate the loss of the corresponding query set $\mathcal{L}_q$ by Eq.\eqref{eq:11}-Eq.\eqref{eq:13} as well. 

\noindent \textbf{Objective and Training Process.}
During the meta training-stage, the objective of \our is to minimize the sum of query loss for all tasks, and the overall loss is:
\begin{equation}\label{eq:17} 
\mathcal{L} = arg\min_\mathbf{\Theta} \sum \mathcal{L}_{q},
\end{equation}

where $\mathbf{\Theta}$ represents all trainable parameters.

\subsection{Algorithm of \our}
We summarize the overall training procedure in Algorithm \ref{alg:cap}.
\begin{algorithm}[t]
\caption{Training framework}
\label{alg:cap}
\textbf{Input}: 
Training tasks $\mathcal{R}_{meta-training}$, initial model parameter $\mathbf{\Theta}$\\
\hspace*{1.722em} Pre-trained KG embedding (excluding relation in $\mathcal{R}_{meta-training}$)
\begin{algorithmic}[1]
\WHILE{not done}
\STATE Sample a task $\mathcal{T}_i = \{\mathcal{S}_i,\mathcal{Q}_i\} $ from $\mathcal{R}_{meta-training}$

\STATE Get $\mathbf{R^s}$ from $\mathcal{S}_i$ by Eq.\eqref{eq:1}-Eq.\eqref{eq:7}
\STATE Get negative sample of $\mathcal{S}_i$ by Eq.\eqref{eq:8}-Eq.\eqref{eq:10}
\STATE Calculate the loss of $\mathcal{S}_i$ by Eq.\eqref{eq:11}-Eq.\eqref{eq:13}
\STATE
Update the embedding of the task-specific relation $\mathbf{R^q}$ with gradient descent by Eq.\eqref{eq:14}
\STATE Get negative samples of $\mathcal{Q}_i$ by Eq.\eqref{eq:8}-Eq.\eqref{eq:10}
\STATE Calculate the loss of $\mathcal{Q}_i$ by Eq.\eqref{eq:11}-Eq.\eqref{eq:13}
\STATE Update whole model parameters $\mathbf{\Theta} \leftarrow \mathbf{\Theta} - \mu\nabla \mathcal{L}$
\ENDWHILE
\end{algorithmic}
\end{algorithm}

\subsection{Difference from RotatE}
RotatE \cite{sun2018rotate} is an embedding-based KGC method that uses a self-adversarial negative sampling technique to effectively optimize the model. Our approach differs from RotatE in a major way: We consider the similarity between the positive triple and negative triple as the weight of each negative triple, but RotatE considers the 
distribution of negative triples and treats the probability as the weight of each negative triple. Therefore, the weights of the negative samples in RotatE are independent of the positive samples. As we will show in the experiments (section 5.5), \our can achieve a better performance than RotatE's self-adversarial negative sampling under the few-shot setting.

\section{Experiments}\label{sec:exper}

\subsection{Datasets and Evaluation Metrics}
We conduct experiments on 
NELL-One and Wiki-One, constructed by \cite{xiong2018one}. In both datasets, relations with more than 50 but less than 500 triples are selected as few-shot relations, and the remaining relations are treated as background relations. We use 51/5/11 and 133/16/34 few-shot relations for training/validation/testing in NELL-One and Wiki-One, respectively. The statistics of both datasets are shown in Table \ref{table:1}.
 
To evaluate the performance of \our and all baselines, we utilize two metrics: mean reciprocal rank (MRR) and Hits@K. MRR is the mean reciprocal rank of correct entities, and Hits@K is the proportion of correct entities ranked in the top $k$.

\begin{table}[t]
\caption{Statistics of the Datasets. Columns 2-7 represent the number of entities, relations, triples, relations in $\mathcal{R}_{meta-training}$, relations in $\mathcal{R}_{meta-validation}$, and relations in 
$\mathcal{R}_{meta-testing}$, respectively.}\label{table:1}
\centering
\begin{tabular}{|c|c|c|c|c|c|c|} 
 \hline
 Dataset & \#Ent & \#Rel & \#Triples & \#Train Rel & \#Valid Rel & \#Test Rel  \\
 \hline
 NELL-One & $68,545$ & $358$ & $181,109$ & $51$ & $5$& $11$ \\
 Wiki-One & $4,838,244$ & $822$ & $5,859,240$  & $133$& $16$& $34$\\
 \hline
\end{tabular}
\vspace{-0.1in}
\end{table}
\subsection{Baseline}

\textbf{Traditional embedding-based methods} aim to learn entity and relation embeddings by modeling relational structure in KG. We consider the following widely used methods as baselines: TransE \cite{bordes2013translating}, DistMult \cite{yang2015embedding}, ComplEx \cite{trouillon2016complex}, SimplE \cite{kazemi2018simple}, and RotatE \cite{sun2018rotate}.
 All these methods require sufficient training triples for each relation and do not use local graph structure to update entity embeddings.

\noindent \textbf{FKGC methods} aim to learn long-tail and unseen relations by utilizing deep neural networks to explore the connection between the support set and the query set. We consider the following models as baselines: GMatching \cite{xiong2018one}, MetaR \cite{chen2019meta}, FSRL \cite{zhang2020few}, FAAN \cite{sheng2020adaptive}, GANA \cite{niu2021relational}, and HiRe \cite{wu2022hierarchical}. We run \our 5 times and report the average results.
\subsection{\our Setups}
The pre-trained entity and relation embeddings are obtained from TransE. The embedding dimension is set to 50 and 100 for NELL-One and Wiki-One, respectively. We use Adam \cite{kingma2015adam} with the initial learning rate of 0.01 to update parameters. The number of negative samples is 5, the margin $\gamma$ is 12.0, the step size $\eta$ is 1, and the number of neighbors is 25 on both datasets. The model with the highest MRR on the validation set is applied as the final model. The optimal hyperparameters are tuned on the validation set by grid search. We conduct \our on a server with a Tesla V100 GPU (32G).

\subsection{Overall Evaluation Results and Analysis}

The performances of all models on NELL-One and
Wiki-One are reported in Table \ref{table:2}. Compared to the traditional embedding-based methods, incorporating graph neighbors is effective for learning entity embedding. \our outperforms the other FKGC models on both datasets. Compared with the runner-up results, the improvements obtained by \our in terms of MRR, Hits@10, Hits@5, and Hits@1 are 4.9\%, 10.2\%, 8.2\%, 2.8\% on NELL-One, and 2.2\%, 2.3\%, 4.3\%, 3.1\% on Wiki-One, respectively. 

\begin{table*}[t]
\caption{Results of 5-shot KGC. \textbf{Bold} numbers represent the best results and \underline{underline} numbers denote the runner-up results. $\dagger$ cites the result from \cite{sheng2020adaptive}, $*$ cites the result from their original papers.}
\label{table:2}
\centering
\resizebox{\textwidth}{!}{%
  \begin{tabular}{l||cccc|cccc}
  \hline
     & \multicolumn{4}{c|}{NELL-One}   & \multicolumn{4}{c}{Wiki-One}  \\
      \cline{2-9}
      Model & {MRR} & {Hits@10} & {Hits@5} & {Hits@1} & {MRR} & {Hits@10} & {Hits@5} & {Hits@1} \\
      \hline
    TransE$^\dagger$ & 0.174 & 0.313 & 0.231 & 0.101 & 0.133 & 0.187 & 0.157 & 0.100 \\
    DistMult$^\dagger$ & 0.200 & 0.311 & 0.251 & 0.137 & 0.071 & 0.151 & 0.099 & 0.024\\
    ComplEx$^\dagger$ & 0.184 & 0.297 & 0.229 & 0.118 & 0.080 & 0.181 & 0.122 & 0.032 \\
    SimplE$^\dagger$ & 0.158 & 0.285 & 0.226 & 0.097 & 0.093 & 0.180 & 0.128 & 0.043 \\
    RotatE$^\dagger$ & 0.176 & 0.329& 0.247 & 0.101 & 0.049 & 0.090 & 0.064 & 0.026 \\
    \hline
    GMatching$^\dagger$ & 0.176 & 0.294 & 0.233 & 0.113 & 0.263 & 0.387 & 0.337 & 0.197 \\
    MetaR$^\dagger$ & 0.209 & 0.355 & 0.280 & 0.141 & 0.323 & 0.418 & 0.385 & 0.270\\
    FSRL$^\dagger$ & 0.153 & 0.319 & 0.212 & 0.073 & 0.158 & 0.287 & 0.206 & 0.097\\
    FAAN$^\dagger$  & 0.279 & 0.428 & 0.364 & 0.200 & 0.341  & 0.463 & 0.395 & 0.281  \\
    GANA$^*$ & \underline{0.344} & 0.517 & 0.437 & \underline{0.246} & 0.351 & 0.446 & 0.407 & 0.299 \\
    HiRe$^*$ & 0.306 & \underline{0.520} & \underline{0.439} & 0.207 & \underline{0.371} & \underline{0.469} & \underline{0.419} & \underline{0.319}\\
    \hline
 \our& \textbf{0.361$\pm$0.011}  & \textbf{0.573$\pm$0.009} & \textbf{0.475$\pm$0.010} & \textbf{0.253$\pm$0.013}  & \textbf{0.379$\pm$0.008}  & \textbf{0.480$\pm$0.012} & \textbf{0.437$\pm$0.008} & \textbf{0.329$\pm$0.011}\\
    \hline
  \end{tabular}
  }
\end{table*}

\subsection{Ablation Study}
\our is composed of two modules, including a dynamic relation-aware entity encoder and negative sampling. To investigate the contributions of each component, we conduct the 5-shot KGC with different settings. The results are summarized in Table \ref{table:3}.

\noindent \textbf{Entity Encoder Variants:} We analyze the impact of the neighboring relation in Eqs.\eqref{eq:3} and \eqref{eq:4} by removing $r_i$ from Eq.\eqref{eq:3} or adding $c_i$ in Eq.\eqref{eq:4}. Besides, we remove the attention mechanism in Eq.\eqref{eq:3}. The results show that neighboring relation and attention mechanism can benefit model performance. It illustrates semantic information of relations can improve the entity representation, and different relations should have different impacts on the entity itself. Since the effect of the attention mechanism depends on neighbors, Wiki-One has much sparser neighbors than NELL-One \cite{niu2021relational}, so the attention mechanism plays a small role in Wiki-One.

\noindent \textbf{Negative Sampling Variants:} To inspect the effectiveness of the negative sampling and attention-based loss functions, we conduct five different experiments. (A) We use only one negative sample in Eq.\eqref{eq:13}. (B) We remove the negative attention mechanism in Eq.\eqref{eq:13}. (C) We remove the candidate pruning stage. (D) We remove the candidate pruning stage and negative attention mechanism. (E) We replace Eq.\eqref{eq:13} with RotatE \cite{sun2018rotate} self-adversarial negative sampling loss.
Experimental results show that the negative sampling strategy plays a key role in the success of \our. 
\begin{table*}[t]
  \caption{Ablation Study}
\label{table:3}
\centering
\resizebox{\textwidth}{!}{%
  \begin{tabular}{l||cccc|cccc}
  \hline
      & \multicolumn{4}{c|}{NELL-One}   & \multicolumn{4}{c}{Wiki-One}\\
      \cline{2-9}
      Model & {MRR} & {Hits@10} & {Hits@5} & {Hits@1} & {MRR} & {Hits@10} & {Hits@5} & {Hits@1} \\
      \hline
      whole model & \textbf{0.372}  & \textbf{0.580} & \textbf{0.477} & \textbf{0.257}  & \textbf{0.387}  & \textbf{0.486} & \textbf{0.443} & \textbf{0.339}\\
      \hline 
    
    Eq.\eqref{eq:3} w/o $r_i$ & 0.339 & 0.535 & 0.427 & 0.222 & 0.362 & 0.468 & 0.410 & 0.299 \\
    Eq.\eqref{eq:4} with $c_i$ & 0.358 & 0.573 & 0.471 & 0.256 & 0.367 & 0.477 & 0.424 & 0.302 \\
    Eq.\eqref{eq:3} w/o $\alpha_i$ & 0.326 & 0.526 & 0.407 & 0.235 & 0.377 & 0.483 & 0.433 & 0.315 \\
    Eq.\eqref{eq:13} with one negative sample& 0.294  & 0.520 & 0.428 & 0.210  & 0.349 & 0.451 & 0.417 & 0.311\\
    Eq.\eqref{eq:13}  w/o negative attention & 0.293 & 0.494 & 0.416 & 0.213 & 0.298 & 0.387 & 0.371 & 0.257\\
    w/o candidate pruning & 0.298 & 0.507 & 0.425 & 0.217 & 0.311 & 0.445 & 0.360 & 0.243\\
    w/o candidate pruning and negative attention & 0.257 & 0.447 & 0.396 & 0.192 & 0.286 & 0.363 & 0.321 & 0.242\\
   RotatE self-adversarial negative sampling & 0.268 & 0.479 & 0.365 & 0.165 & 0.310 & 0.389 & 0.401 & 0.255\\
    \hline
  \end{tabular}
  }
\end{table*}

\subsection{Influence of Size of Few-shot Support Set and Negative Sample}
\begin{figure}[t]
    \centering
    \subfloat[\centering\label{fig:2(a)} ]{{\includegraphics[width=5.7cm]{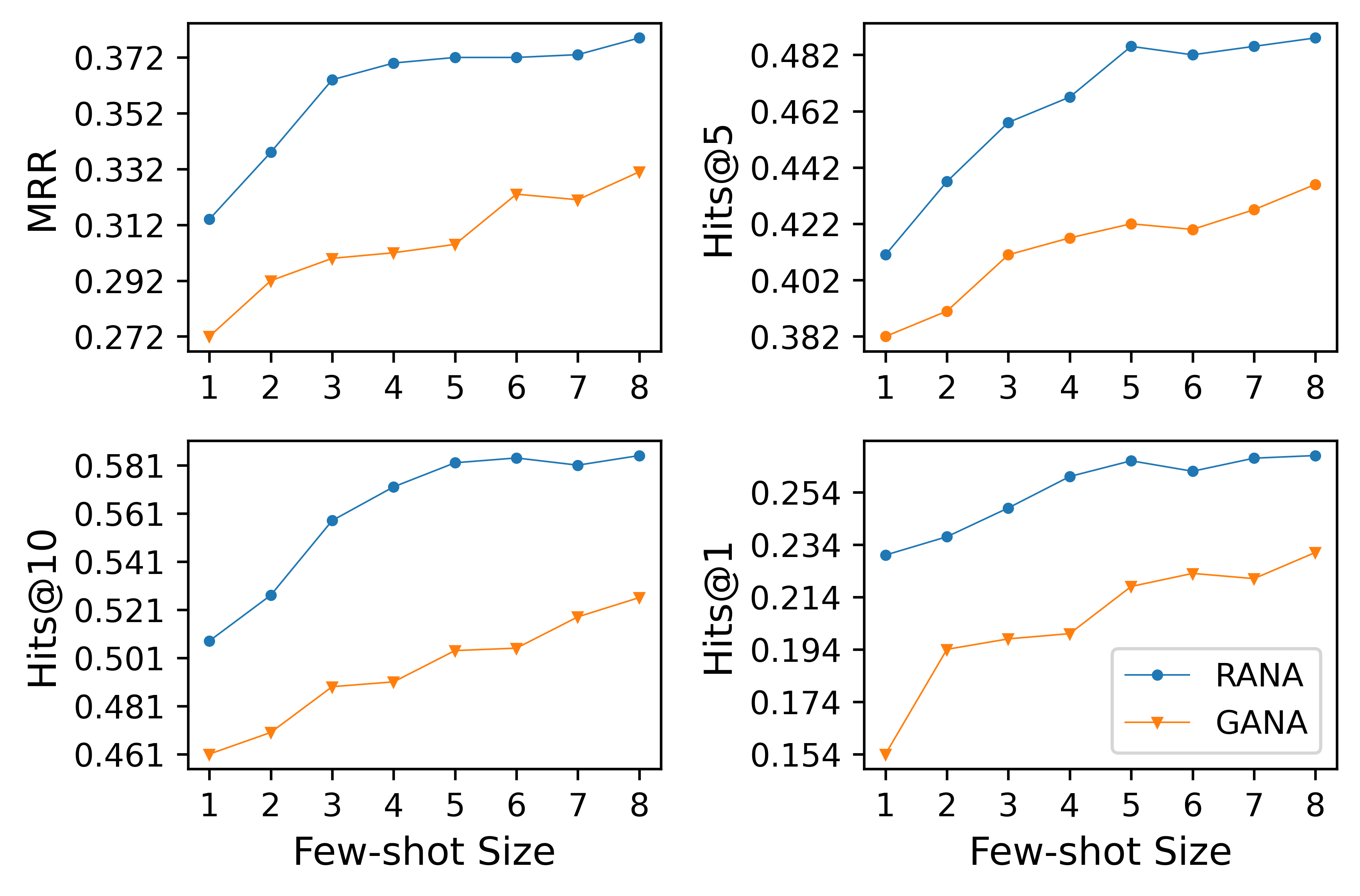} }}%
    \qquad
    \subfloat[\centering\label{fig:2(b)} ]{{\includegraphics[width=5.7cm]{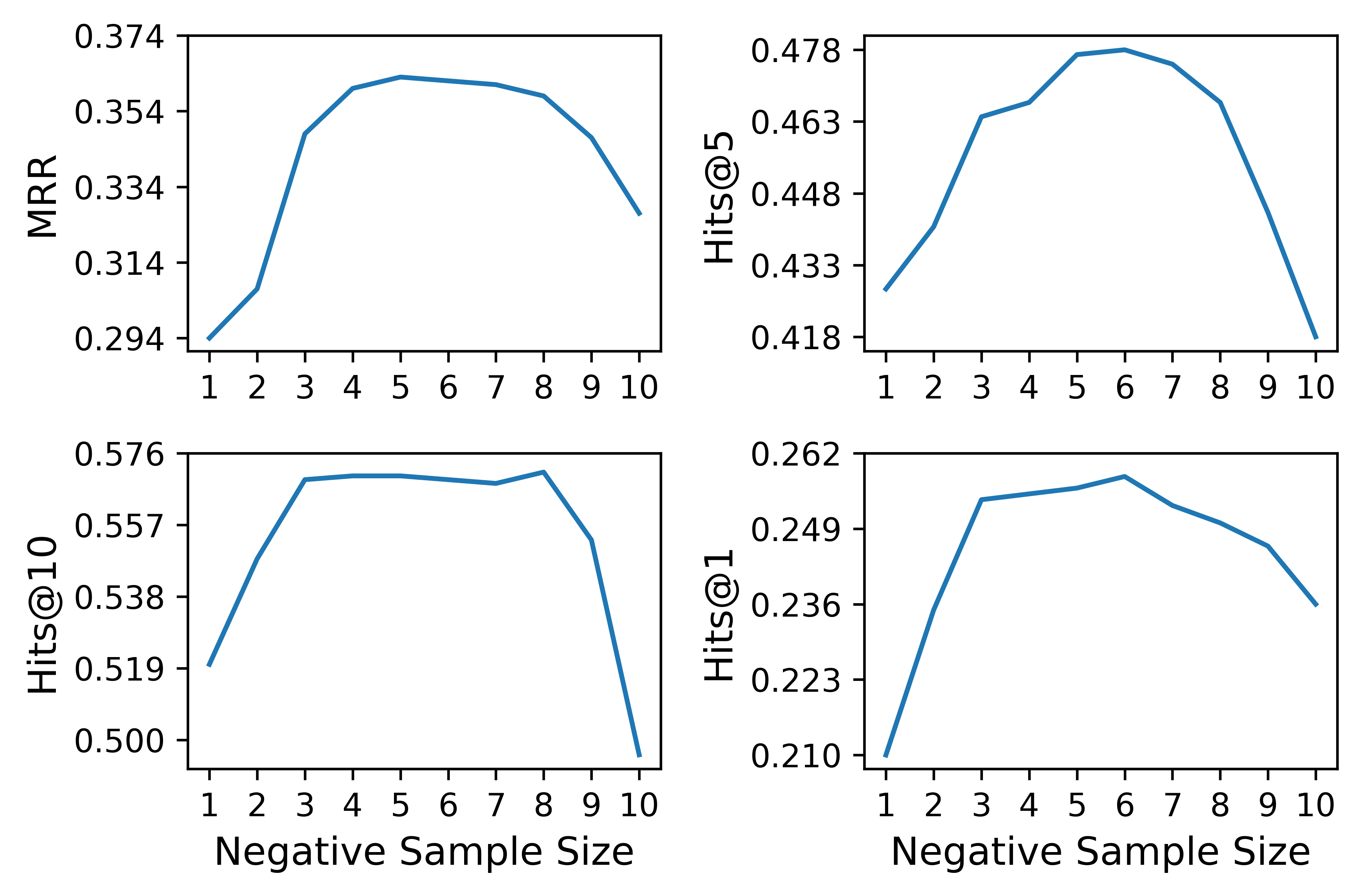} }}%
    \caption{(a) Influence of Few-shot Support Set Size,(b) Influence of Negative Sample Size}%
    \label{fig:2}%
\end{figure}

To analyze the impact of support set size, we compare \our with GANA on NELL-one. Fig.\ref{fig:2(a)} shows the performances with support set size from 1 to 8.
\our outperforms GANA under different sizes of support sets, showing the effectiveness of \our. After the 5-shot, the improvement of \our is not significant. We randomly select 20 facts from the relation \textit{teamcoach} to analyze the errors in the 5-shot setting. \our predicts 12 out of 20 true tail entities in top 10. Among the other 8 facts, 4 of them have incorrect ground truth tail entities, and 3 of them have neighbors fewer than 10. For these cases, increasing the size of the support set is unlikely to change the results.

We conduct an experiment to analyze the influence of the negative sample size. Fig.\ref{fig:2(b)} shows the performance of \our on NELL-One with the negative sample size from 1 to 10.
The performance improves initially when increasing the negative sample size.
After size 6, the performance begins to drop due to the class imbalance issue. Empirically, we recommend a negative sample size of 3 to 5.

\section{Conclusion}\label{sec:conclusion}
In this paper, we propose a relation-aware network with attention-based loss for FKGC tasks. We strategically select multiple negative samples instead of one and propose an attention-based loss to differentiate the importance of each negative sample. A dynamic relation-aware entity encoder is designed to learn a context-dependent entity representation. The experimental results demonstrate that \our outperforms other SOTA baselines on two benchmark datasets.

\bibliographystyle{splncs04}
\bibliography{reference}
%




\end{document}